\documentclass[11pt,letterpaper]{article}
\usepackage{emnlp2017}

\usepackage{url}
\usepackage[pdftex]{graphicx}
\usepackage{booktabs}
\usepackage{latexsym}
\usepackage{amsmath}
\usepackage{amsfonts}
\usepackage{multirow}
\usepackage{enumitem}
\usepackage{color}
\usepackage{float}
\usepackage{listings}
\usepackage{anyfontsize}
\usepackage{pifont}
\usepackage{caption}
\usepackage{pbox}
\usepackage{xspace}
 
\usepackage{microtype} 
\makeatletter

\usepackage{booktabs,arydshln}
\makeatletter
\def\adl@drawiv#1#2#3{%
        \hskip.5\tabcolsep
        \xleaders#3{#2.5\@tempdimb #1{1}#2.5\@tempdimb}%
                #2\z@ plus1fil minus1fil\relax
        \hskip.5\tabcolsep}
\newcommand{\cdashlinelr}[1]{%
  \noalign{\vskip\aboverulesep
           \global\let\@dashdrawstore\adl@draw
           \global\let\adl@draw\adl@drawiv}
  \cdashline{#1}
  \noalign{\global\let\adl@draw\@dashdrawstore
           \vskip\belowrulesep}}
\makeatother

\usepackage{times}
\usepackage{latexsym}

\emnlpfinalcopy

\def\emnlppaperid{***}

\newcommand{\systemx}{{\sc SwellShark}\xspace}

\newcommand{\squishlist}{
 \begin{list}{$\bullet$}
  { \setlength{\itemsep}{0pt}
    \setlength{\parsep}{2pt}
    \setlength{\topsep}{2pt}
    \setlength{\partopsep}{0pt}
  }
}
\newcommand{\squishend}{\end{list}}

\title{\systemx: A Generative Model for Biomedical \\ Named Entity Recognition without Labeled Data }

\author{Jason Fries, Sen Wu, Alex Ratner, Christopher R\'e \\
  Stanford University / Stanford, CA \\
  {\tt \{jfries,senwu,ajratner,chrismre\}@cs.stanford.edu} \\}

\date{}

\begin{document}

\maketitle

\begin{abstract}

We present \systemx, a framework for building biomedical named entity recognition (NER) systems quickly and without hand-labeled data. Our approach views biomedical resources like lexicons as function primitives for autogenerating \emph{weak supervision}. We then use a generative model to unify and denoise this supervision and construct large-scale, probabilistically labeled datasets for training high-accuracy NER taggers. In three biomedical NER tasks, \systemx achieves competitive scores with state-of-the-art supervised benchmarks using no hand-labeled training data. In a drug name extraction task using patient medical records, one domain expert using \systemx achieved within 5.1\% of a crowdsourced annotation approach -- which originally utilized 20 teams over the course of several weeks -- in 24 hours. 



\end{abstract}


%
%
\section{Introduction}

Named-entity recognition (NER) is a foundational NLP task that is traditionally approached as a 
supervised learning problem. In this setting, state-of-the-art NER systems often require 
considerable manual feature engineering to learn robust models using hand-labeled training data. 
Recent success in deep learning for NER \cite{lample2016neural} suggests that automatic feature 
extraction will largely replace this process. However, this shifts the burden to constructing 
the massive hand-labeled training sets needed for robust deep models. 

How do we obtain enough training data to fit these complex models? Crowdsourcing offers one 
way of generating large-scale labeled data, but the process is expensive, especially when 
annotators require specialized domain knowledge or data has privacy concerns preventing 
distribution \cite{sabou2012crowdsourcing,gokhale2014corleone}. Furthermore, even expert inter-annotator agreement 
rates can be low for certain tasks. 

In NLP, another common approach is \emph{distant supervision} \cite{mintz2009distant} where 
structured resources like ontologies and knowledge bases are used to heuristically label 
training data. While noisy, this technique has shown empirical success. Distant supervision 
is commonly used with a few, canonical structured resources like Freebase \cite{bollacker2008freebase}, 
weighting each resource equally when labeling data. However, in biomedicine we are faced with a 
wide space of curated resources; NCBO Bioportal \cite{whetzel2011bioportal} currently houses 
541 distinct biomedical ontologies. These resources contain different hierarchical structures, 
concept granularities, and otherwise overlap or conflict in their definitions of 8 million 
entities. Any single ontology may have widely varying accuracy depending on the target task, making them
difficult to combine using simple methods like majority vote.  

We present \systemx, a framework for quickly building biomedical NER taggers using lexicons, heuristics, and other forms of \emph{weak supervision} instead of hand-labeled data. Our approach effectively subsumes both crowdsourcing and distant supervision,
automatically modeling all such inputs as a generative labeling process. This functional view allows us to take advantage of recent advances in denoising weak supervision, more effectively unifying large-scale biomedical resources. 

Our approach greatly simplifies supervision in an NER system. Traditional distant supervision pipelines 
consist of three basic components: (1) \emph{candidate generation}, i.e., identifying potential 
entities for labeling and classification; (2) \emph{labeling heuristics} for generating noisy 
labels; and (3) \emph{features} describing candidates for classification. Previously, 
each of these has required human engineering or supervision. In \systemx we show that 
in the presence of structured resources like lexicons and ontologies, components (1) and (2) can be largely automated. 
When coupled with automatic feature extraction models like LSTMs, we create a powerful end-to-end 
pipeline that requires dramatically less human input and can train high-accuracy taggers 
using \emph{unlabeled data}.

The central argument of this work is that modeling noise in supervision resources is such a powerful strategy that it
enables tremendous performance gains from even simple NER techniques. This allows us to focus exclusively on the resources used to supervise a model instead of diffusing human effort across an entire extraction pipeline.

Our three core contributions are summarized as: 

\squishlist
\item {\bf Automatic Candidate Generation:}
\systemx automatically generates potential or \emph{candidate} entity mentions in documents, a heuristic process that traditionally required non-trivial engineering. Since candidates define the space over which we both provide supervision and learn, selecting the right approach is critical to overall performance. 

\item {\bf Autogenerated Supervision:} 
\systemx only requires a set of positive and negative lexicons as baseline input. Several classes of automatic supervision generators apply transformations to these lexicons and efficiently generate a large space of noisy supervision with minimal human input.

\item {\bf Weakly-supervising Sequences:} 
\systemx allows us to ``compile" supervision inputs, like lexicons and other heuristic rules, directly into a sequence prediction model. We propose a multinomial generative model to explicitly learn entity boundaries, a key part of NER, and model the accuracies of our underlying supervision sources.  
\squishend

Modeling noise while generating data is critical for scaling, where we improve tagger accuracy by using more unlabeled data to train our models. With \systemx, we construct weakly-labeled training sets of up to 100K documents, providing boosts of up to 6.7\% (4.9 F1 points) over the same models trained on small (	$\leq$ 1K) document collections. With scaling, we can achieve competitive results to state-of-the-art supervised models.

Finally, as an applied validation challenge, we used \systemx to build an NER system in another biomedical domain: tagging drug names in clinical discharge summaries. We report the performance achieved by a single domain expert given 24 hours for development and model training. \systemx scored within 5.1\% of a crowdsourced annotation approach, which originally utilized 20 teams over the course of several weeks. 

%
%
\section{Related Work}
\label{sec:related}

Domain-specific NER tasks are a well-studied NLP problem. The best performing systems use supervised or semi-supervised learning and require hand-annotated training data. In biomedical NER, supervised methods 
using CRFs are the standard \cite{settles2004biomedical,leaman2008banner} though RNNs/LSTMs are increasingly common \cite{sahu2016recurrent,dernoncourt2016identification}. Semi-supervised methods that augment labeled datasets with word embeddings \cite{tang2014evaluating,kuksa2010semi} or bootstrapping techniques \cite{vlachos2006bootstrapping} have been shown to outperform supervised baselines in tasks like gene name recognition. Unlike these existing approaches, \systemx does not require hand-labeled training data and is agnostic to the choice of discriminative model.


Leveraging existing resources to heuristically label data has received considerable research interest. 
Distant supervision \cite{craven1999constructing,mintz2009distant} uses knowledge bases to supervise relation extraction tasks.  Recent methods incorporate more generalized knowledge into extraction systems. Natarajan et al.\shortcite{natarajan2016deep} used Markov Logic Networks to encode commonsense domain knowledge like ``home teams are more likely to win a game" and generate weak training examples. \systemx is informed by these methods, but uses a generative model to unify and model noise across different supervision sources.

%
%

\section{Background}

\paragraph{Biomedical NER}

Identifying named entities is a core component of applied biomedical information extraction systems and a critical subtask in
\emph{normalization}, where entities are mapped to canonical identifiers, and \emph{relation extraction}, where we identify $n$-arity semantic connections between entities. Ontologies are key artifacts in formalizing biological concepts for normalization and use in computational systems.  Biomedical NER focuses on identifying these concepts. For example we would label a sentence as: ``\textbf{\textit{Primary pulmonary hypertension}} \emph{is a rare, progressive and incurable disease.}" to identify a disease name. 

For simplicity, in this work we assume each entity is a binary classification task, i.e., each tagger predicts one entity type, although our method generalizes to multi-class settings without extensive changes. We focus on the recognition part of named entity extraction and do not address normalization.

\label{sec:dp}
\paragraph{Data programming:}
Ratner et al. \shortcite{ratner2016data} proposed \emph{data programming} as a method for programmatic training set creation. 
In data programming, a collection of user-provided rules called \emph{labeling functions} (LFs) are modeled as a generative process of training set labeling. Labeling functions may overlap and conflict in their labels, as long as the majority have accuracies greater than 50\%. By default, it's assumed that labeling functions are conditionally independent. Fitting a generative model allows us to automatically estimate these accuracies without ground truth data. The resulting model is then used to construct large-scale training sets with probabilistic labels.  

Formally, labeling functions are black box functions which label some subset of data. In our setting, given a set of candidates for a single entity class (e.g., disease names) and corresponding binary labels, 
$(x,y) \in \mathcal{X}\times\{-1,1\}$, where the $y$ are unseen, a labeling function $\lambda_i$ maps:
\vspace{-0.05in}
\begin{align*}
\lambda_i : \mathcal{X} \mapsto \{-1,0,1\}
\end{align*}
where 0 means a labeling function abstains from providing a label.
The output of a set of $M$ labeling functions applied to $N$ candidates
is a matrix $\Lambda \in \{-1,0,1\}^{N\times M}$.
Each labeling function is represented by a single accuracy parameter, learned based on observing the agreements and disagreements between overlapping labeling functions.


Instead of binary ground-truth training labels $(x,y)$ with $y\in\{-1,1\}$,
we now utilize the marginal probabilities of our generative model,
$P_{\mu}(Y\ |\ \Lambda) \in [0,1]$, as training labels for a discriminative model, such as logistic regression or an LSTM.
This requires using a \textit{noise-aware} version of our loss function. This can be implemented analytically in the discriminative model or simulated by creating a sampled dataset based on our marginal probabilities.

\section{Methods}
\label{sec:sys}

\begin{figure}[ht!]
\centering
\includegraphics[width=79mm]{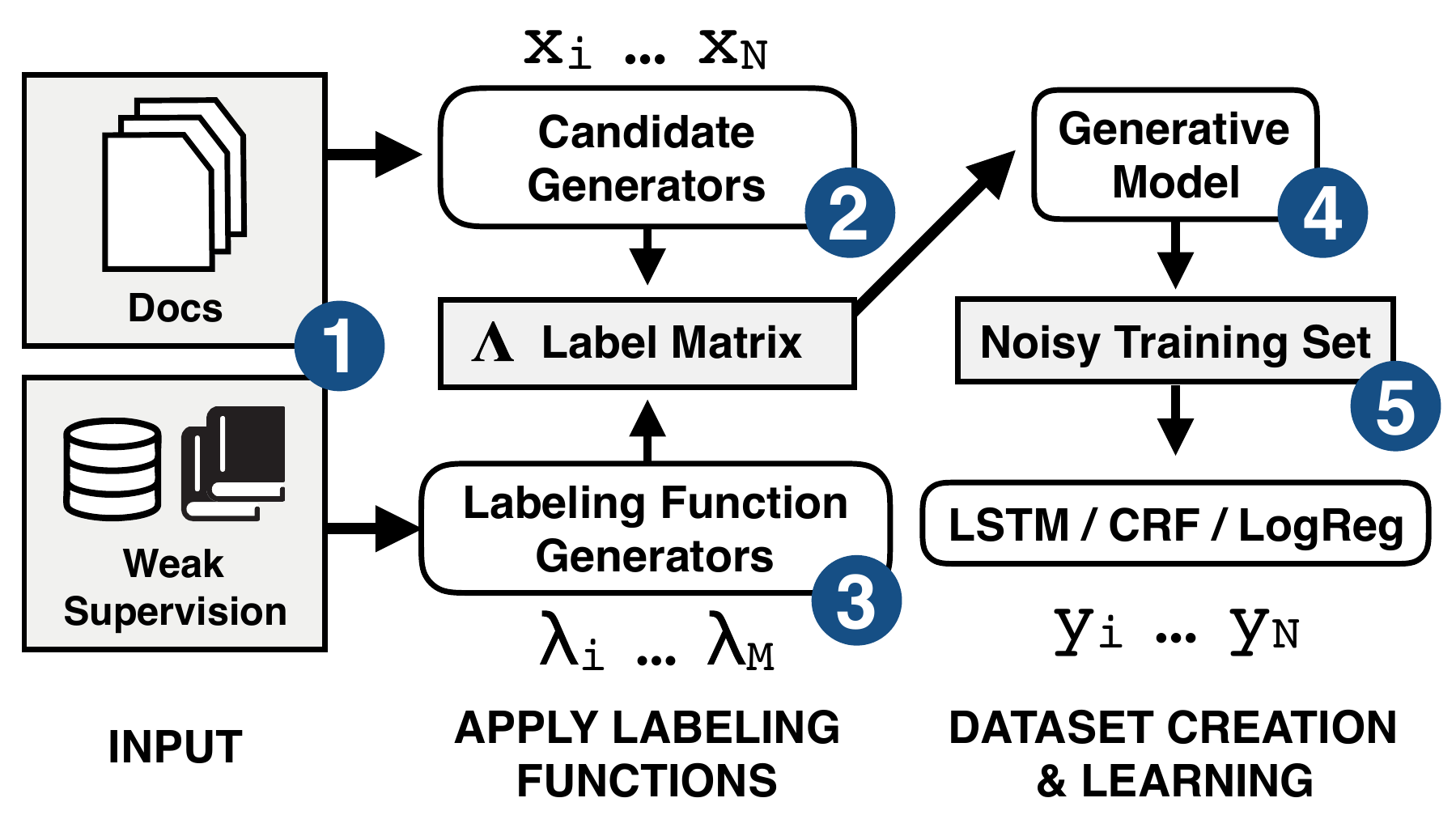} 
\caption{\systemx pipeline. The numbers correspond to the stages described in \S \ref{sec:sys-input} - \S \ref{sec:sys-building}.}
\label{fig:workflow}
\end{figure}

The \systemx pipeline is outlined in Figure \ref{fig:workflow} and consists of the following stages: 1) providing unlabeled documents and defining weak supervision input; 2) using generators to transform documents into a set of candidates for classification; 3) autogenerating labeling functions using structured resources or user heuristics; 4) fitting a multinomial generative model using the output of all labeling functions as applied to candidates; and 5) generating probabilistically labeled data, which can then be used with any off-the-shelf classification model. Details for each stage are described below.

\subsection{SwellShark Input}
\label{sec:sys-input}
\systemx requires as input a collection of unlabeled documents and some form of weak supervision. This is typically a collection of  lexicons, ontologies, and optional heuristic rules. Supervision largely consists of specifying positive and negative lexicons. As a toy example, a minimal drug tagger specification could be ({\tt1}: \emph{antibotic}, {\tt-1}: \emph{amino acid, peptide, or protein}, \emph{gene or genome}), with each semantic category mapping to source lexicons in the Unified Medical Language System (UMLS) \cite{bodenreider2004unified} or other external dictionaries. 

\subsection{Candidate Generators}
\label{sec:sys-ce}
Our approach requires first identifying a set of potential or \emph{candidate} mentions in documents. We define a \textit{candidate generator}, $\Gamma_\phi$, as a function that transforms a document collection $D$ into a candidate set: $\Gamma_\phi: D \mapsto \{x_1,...,x_N\}$.  Each candidate $x$ is defined as a character-level span within a document sentence. Candidate generators are heuristics that can be restrictive e.g., the set of all dictionary matches, or permissive, such as all overlapping $k$-gram spans. 
The choice of heuristic impacts overall performance, since candidates define the space over which we both provide supervision and learn.
We explore the following simple automated generators:


\squishlist
\item {\bf Noun Phrases} All noun phrases as matched using regular expressions over POS tags. 
\item {\bf Dictionary } Domain dictionaries with no domain-specific stopword lists or other lexical curation. 
\squishend

Each heuristic emits $k$-grams candidates; in our experiments $k$ = [1-10]. 
Choosing a heuristic involves some trade-off between development time and performance. 
Hand-tuned matchers require more engineering, but generate more accurate candidate sets.
This requires less weak supervision to train the discriminative model, since the candidate 
generation step acts as an implicit hard filter. In contrast, dictionary and noun phrase 
candidates are generated automatically, but create additional challenges during learning. 
Dictionary matches limit recall, impacting generalizability; noun phrase candidates 
generate larger sets, introducing more noise during labeling function application.

\subsection{Labeling Function Generators}
\label{sec:sys-lfs}

Labeling functions are a generalization of strategies used in distant supervision. For example, in disease name tagging we can define a labeling function that outputs {\tt 1} if a candidate occurs in a \emph{disease or syndrome} lexicon and another function which outputs {\tt -1} if it's found in a \emph{gene or genome} lexicon.
Several examples are shown in Figure \ref{fig:lfs}.

\begin{figure}[ht]
\begin{center}
\advance\leftskip-3cm
\advance\rightskip-3cm
\fontsize{9.5}{10.5}
\begin{verbatim}
def LF_in_lexicon(c):
   t = c.text()
   return 1 if t in umls_disease else 0
 
def LF_idf_filter(c):
   return -1 if idf(c) <= 4.0 else 0
    
def LF_temporal_modifiers(c):
   head = c.tokens("words")[0]
   return -1 if head in temp_mod else 0
\end{verbatim}
\caption{Example labeling functions : (top) tests for membership in a lexicon; (middle) filters by candidate inverse document frequency; and (bottom) defines a compositional grammar rule for rejecting candidates beginning with a temporal modifier, e.g., \emph{*recurrent* carcinoma}, \emph{*childhood* cancer}.}
\label{fig:lfs}
\end{center}\end{figure}

\begin{figure*}[htbp!]
\centering
\includegraphics[width=150mm]{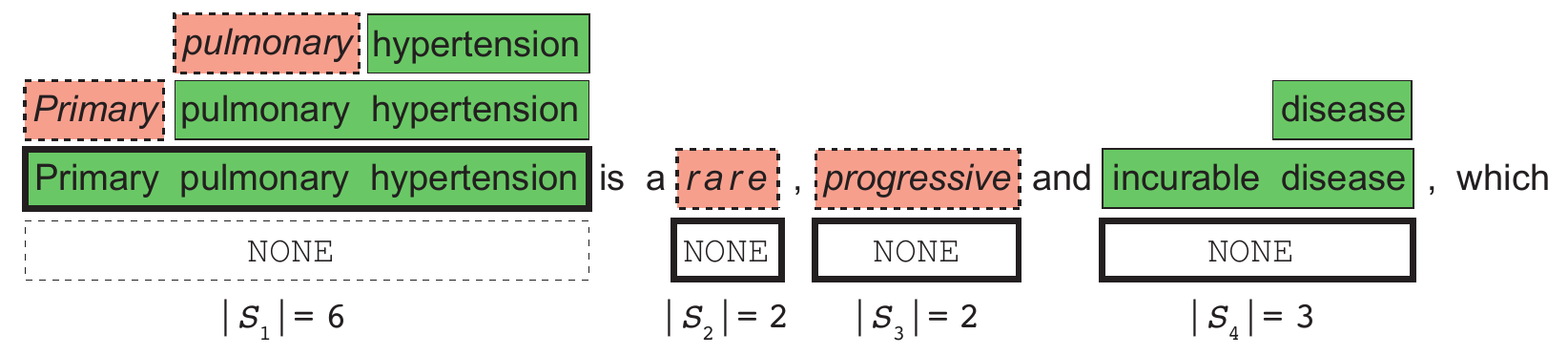}
\caption{Sentence partitioning example. Each overlapping candidate set defines a \emph{spanset} $s_i$. 
Given one positive and negative LF, green is positive, red is negative, and bold outlines indicate the true mention. {\tt NONE} is the absence of an entity. Overlapping candidates, as in $s_1$, cause errors in probability estimation. }
\label{fig:sentence-partition}
\end{figure*}


Labeling functions using structured resources assume predictable froms, meaning most lexical-based supervision can be autogenerated. We define a \textit{labeling function generator} (LFG), $\Gamma_\lambda$, as a function which accepts a weak supervision resource, $R$, e.g., an ontology or empirical attributes like term frequencies, and automatically generates one or more labeling functions, $\Gamma_\lambda : R \mapsto  \{\lambda_1,...,\lambda_N\}$.
LFGs automate a large space of supervision from minimal manual input, such as a collection of positive and negative lexicons. We implemented LFGs as described in Table \ref{tbl:lf-generators}.

\begin{table}[ht!]
\centering
\begin{tabular}{l|l}
\toprule
\bf Type & \bf LF Generator Description \\ \hline 
Lexical & $\lambda \mapsto$ $c_i$ $\in$ $lexicon$ \\\cline{2-2}
        & $\lambda \mapsto$ {\tt TailWord}($c_i$) $\in$ $lexicon$ \\\cline{2-2}
        & $\lambda \mapsto$ {\tt AbbrvDef}($c_i$) $\in$ $lexicon$\\
        & \ \ \ $c_i$ is a parenthetical def. e.g., \\ 
        & \ \ \ ``Myotonic dystrophy (DM)" \\ \hline
Filters & $\lambda \mapsto$ {\tt IDF} ($c_i$) $>$ $threshold$ \\\cline{2-2}
        & $\lambda \mapsto$ {\tt DF} ($c_i$) $>$ $threshold$ \\\cline{2-2}
        & $\lambda \mapsto$ {\tt PhraseFragment}($c_i$) \\
        & \ \ tail word is an adjective  \\ \hline
LF  & $\lambda_a(\lambda_b)$ $\mapsto$ {\tt Children}($c_i$) \\   
Modifiers         & \ \ \ cascade $\lambda_b$ to all fully \\ 
         & \ \ \ nested child candidates  \\\cline{2-2}
         &  $\lambda_a(\lambda_b)$ $\mapsto$  $\lambda_a(c_i) \circ \lambda_b(c_i)$ \\  
         & \ \ \ logical composition of LFs \\      
\bottomrule 
\end{tabular}
\caption{Labeling function generators.}
\label{tbl:lf-generators}
\end{table}

A single generator can instantiate multiple positive and negative labeling functions. This occurs in the case of overlapping candidate spans, where we allow supervision to cascade to nested child entities. In Figure \ref{fig:sentence-partition} some children of ``primary pulmonary hypertension" are dictionary members and are assigned positive labels. However, if we wish to enforce longest match, we can define a dictionary generator that votes positive on the longest match and negative on all child nodes. Other complex primitives are possible using compositions, such as synonym or negation detection. Our goal is to make it easy and fast to use combinations of LFGs and LFs to provide supervision for a new tagger.

\subsection{Multinomial Generative Model}
\label{sec:sys-learning}

A key challenge of our simple, unified framework for accepting weak supervision is that it involves overlapping candidates. 
The generative model originally proposed by data programming assumes candidates are independent, and does not account for dependencies induced by overlapping spans. Figure \ref{fig:sentence-partition} shows how this can lead to incorrect marginal probability estimates. In this example, all variants of ``hypertension" are found in a single lexicon, overestimating positive label probabilities for nested candidates. To address this bias, we extend the generative model to learn mutual exclusion constraints. For NER, this change is vital because it allows us to learn entity boundaries while maintaining simple labeling function semantics, i.e., voting on discrete candidates. In our experiments, modeling these dependencies improves overall F1 score by up-to 4.7\% (3.4 points).

We define a \emph{spanset}, $s$, as a collection of overlapping candidates within a sentence as partitioned by the labeling matrix $\Lambda$, where each candidate's row contains the labels generated by all labeling function. Candidates with $\ge1$ positive, overlapping labels form the basis of each spanset. Candidates with $\ge1$ negative labels and 0 positives are added if such addition does not join two disconnected spansets. Unlabeled candidates are removed.   

Each spanset is represented as a matrix $\hat X \in \mathbb{R}^{M\times K}$ where $M$ is the number of labeling functions and $K$ is the number of overlapping candidates per spanset plus the {\tt NONE} class. In our datasets, these matrices are column sparse with small $K$. Pathologically long cases are filtered out. 

Positive label semantics remain unchanged in this framing. However negative labels are now underspecified since they don't provide enough supervision to perfectly map to a sequence prediction task. A candidate may be completely wrong (the {\tt NONE} class) or a partial match. For example, in Figure \ref{fig:sentence-partition} a negative vote for ``primary" doesn't tell us if the whole spanset is wrong or if it's a subset of a correct entity (as is the case).  Our convention is if the $i$th LF votes negatively on a candidate $j$, this is expressed as having uniform distribution over all $K$ columns except $j$ in row i of $\hat X$.

Learning is otherwise unchanged from the binary case, except we now apply the softmax function to each spanset matrix $\hat X$ to compute marginals. 
\begin{center}
$P(Y=j | \hat X; w) = \frac{exp({w^T \hat X_j})}{\sum_{k=1}^K exp(w^T \hat X_k)}$
\end{center}

\subsection{Sampling for Dataset Construction}
\label{sec:sys-building}
After training, each spanset now defines a multinomial distribution
$s_i$ = \{($c^{i}_{j}$, $p^{i}_{j}$); $j=1 ... k$\}, where $p$ is the probability of each candidate within the spanset. We treat all spansets in a sentence, $\mathbb{S}$, as as sampling distribution $\mathbb{S}$ $\sim$ $s_1 \times s_{2} \times ... s_N$
to generate noisily labeled tag sequences. We assume spansets are independent and generate 10 samples per observed sentence, sampling once per spanset.

%
%
\section{Experiments}


\label{sec:exp}

\subsection{Weakly-supervised Taggers}

As our experimental testbed, we built three biomedical taggers: two for disease names (``osteoarthritis'') and one for chemical names (``bupropion hydrochloride''), with each tagger trained using a CRF or LSTM. Each model configuration is evaluated using 25K to 100K unlabeled training documents. 

Our primary experimental questions are: 1) what are the performance trade-offs of different candidate generation heuristics; 2) how well does autogenerated supervision perform; and 3) how quickly can we write a tagger for use in other domains? 

\paragraph{Comparison Systems:} For our baseline comparison system, we use reported benchmarks from TaggerOne \cite{leaman2016taggerone} a state-of-the-art general purpose biomedical NER tagger. Their approach uses a CRF with manually engineered features. 
We compute two simple baselines: a string matching score using all positive lexicons (LEX) and a domain-agnostic stop word list; and the majority vote (MV) across all labeling functions. 

\paragraph{1) Tuning Candidate Generation:} We explored the trade-offs of two automated candidate generation methods compared to a manually-engineered baseline. For each tagger, we implemented an optimized, hand-tuned generator using regular expressions, dictionaries, and other fuzzy matching heuristics. These generators are task-specific, carefully maximizing recall while minimizing false positives. We compute precision-recall curves for each heuristic and report F1 scores when those methods are scaled. We empirically evaluated precision/recall tradeoffs of different $k$-gram token lengths and use $k$=6 for all candidates. Here we use the CDR disease task as our motivating example. 

\paragraph{2) Autogenerating Supervision:} We trained models using only autogenerated, lexicon-based labeling functions and \emph{NounPhrase} candidate generators. Here supervision consists of specifying positive and negative semantic categories selected from the 133 semantic definitions provided by the UMLS.  Lexicons provide a strong baseline supervision framework, but they are usually incapable of modeling the entire dataset generation process. The definition of ground truth often depends on dataset-specific annotation guidelines. For example, in our clinical notes task, mentions of drug names that are negated or describe patient allergies are not considered true mentions; a lexicon alone cannot encode this form of supervision. Choosing what affixes, modifiers, and prepositional phrases constitute an entity can vary greatly across datasets. 

One advantage of \systemx is that this type of dataset-specific supervision is easily introduced into our system by adding more labeling functions. For each PubMed tagger, we extended our lexical supervision with labeling functions to capture annotation guidelines and other dataset-specific supervision. This required writing 20-50 additional labeling functions per task. 

\paragraph{3) Building a Tagger in 24-hours:} We tested out ability to quickly build a high-performance tagger in another biomedical domain. Given a time budget of 24 hours, we used our tools to build a drug name tagger for clinical discharge summaries.  For the autogenerated LF model, our positive supervision consisted of 13 chemical semantic types from the UMLS. This setup required about one hour. We spent the remaining time examining training data to identify syntactical patterns and rules for labeling functions. Due to time constraints and the lack of additional unlabeled data, we did not explore scale-up performance.

\paragraph{Labeling Function Development:} All labeling functions and domain stop word lists were developed iteratively by inspecting unlabeled training documents. Manually written LFs were refined based on empirical accuracy scores from a small set of held-out labeled training instances. No ground truth labels were used to fit our final discriminative models; we only use noise-aware labels produced by the generative model. 


\begin{table*}[htbp!]
\centering
\begin{tabular}{llccc|ccc|ccc}
\toprule
\multirow{2}*{\bf System} & \multirow{2}*{\bf \emph{CandGen}} & \multicolumn{3}{c}{\bf NCBI (Disease)} & \multicolumn{3}{c}{\bf CDR (Disease)} & \multicolumn{3}{c}{\bf CDR (Chemical)} \\ \cmidrule{3-11}
         &         & \bf P     & \bf R     & \bf F1    & \bf P     & \bf R     & \bf F1    & \bf P     & \bf R     & \bf F1    \\
\midrule
TaggerOne$^*$ &            & 83.5 & 79.6 & \bf 81.5 & 83.1 & 76.4 & 79.6 & \bf 92.4 & 84.7 & \bf 88.4 \\ \midrule

Majority Vote & \emph{Hand-tuned}  & \bf 84.5 & 75.5 & 79.8  &  \bf 85.4 & 67.6 & 75.5  & 89.8 & 83.1 & 86.3 \\  
\ding{169} CRF/emb & \emph{Hand-tuned}        & 78.7 & 78.0 & 78.4 & 83.1 & 77.2 & \bf 80.1 & 89.6 & 84.0 & 86.7  \\
\ding{169} LSTM-CRF/emb & \emph{Hand-tuned}   & 81.6 & \bf 80.1  & 80.8  & 81.6 & \bf 78.6 & \bf 80.1  & 89.8 & 85.5 & 87.6 \\ \midrule 

Majority Vote & \emph{Noun Phrase}   & 84.4 & 51.7 & 64.1 & 76.4 & 67.3 & 71.5     & 86.2 & 86.1 & 86.2   \\ 
\ding{169} CRF/emb & \emph{Noun Phrase}    &  56.5 & 64.4 & 60.2 &  81.5 & 75.81 & 78.5 & 89.2 & 86.7 & 87.9 \\ 
\ding{169} LSTM-CRF/emb & \emph{Noun Phrase}    & 64.7 & 69.7 & 67.1 & 80.7 & 77.6 & 79.1 & 88.3 & \bf 88.3 & 88.3  \\ \midrule

Lexicon Baseline  &              & 36.3 & 77.6 & 49.5    & 40.8 & 77.1 & 53.3     & 62.0 & 81.2 & 70.3   \\
\bottomrule 
\end{tabular}
\caption{Best \systemx results compared against supervised baselines. Here we add additional labeling functions to improve performance and report scores for both hand-tuned and automatic candidate generators. 
 \ding{169} indicates the highest scoring model after scaling with additional ($\leq$ 100K) unlabeled documents. * \emph{TaggerOne scores are for NER only.}
}
\label{tbl:dp1}
\end{table*}

\subsection{Materials and Setup}

\paragraph{Datasets:} We evaluate performance on three bioinformatics datasets: (1) the NCBI Disease corpus \cite{dougan2014ncbi}; (2) the BioCreative V Chemical Disease Relation task (CDR) corpus \cite{wei2015overview}; and the i2b2-2009 Medication Extraction Challenge dataset \cite{uzuner2010extracting}. NCBI Disease contains 792 PubMed abstracts separated into training, development, and test subsets ($n$=592/100/100); CDR contains 1,500 PubMed abstracts ($n$=500/500/500); and i2b2-2009  contains 1249 electronic health record (EHR) discharge summaries ($n$=1000/124/125). PubMed datasets are annotated with mention-level disease and chemical entities and i2b2 data with drug names.

For unlabeled PubMed data, we use a 100K document sample chosen uniformly at random from the BioASQ Task 4a challenge \cite{tsatsaronis2015overview} dataset. All LSTM experiments use the 200-dim word2vec embeddings provided as part of this dataset. For clinical text embeddings, we generated embeddings using 2.4M clinical narratives made available as part of the MIMIC-III critical care database \cite{johnson2016mimic}.

Ontologies used in this work include those provided as part of the 2014AB release of the UMLS, and various other disease/chemical ontologies \cite{schriml2012disease,davis2015comparative,rath2012representation,kohler2013human}.

\paragraph{Data Preprocessing:} All datasets are preprocessed using Stanford CoreNLP\footnote{http://stanfordnlp.github.io/CoreNLP/} with default English models for tokenization, sentence boundary detection, POS tagging, and dependency parsing.

\paragraph{Discriminative Models:} We use two external sequence models: CRFsuite \cite{CRFsuite} and a bidirectional LSTM-CRF hybrid \cite{lample2016neural} which makes use of both word and character-level embeddings. Our LSTM-CRF uses automatic feature extraction based on word-level input. CRFs use a generic feature library (see \S Appendix) based on TaggerOne.

\paragraph{Evaluation Measures:} Precision (P), Recall (R), F1-score (F1).


%
%
\section{Results / Discussion}

\begin{table*}[htbp!]
\centering
\setlength\tabcolsep{5pt} 
\begin{tabular}{lccccccc|cccc}
\toprule
\multirow{2}*{\bf Dataset (Entity)} & \multirow{2}*{\bf LFs} &  \multirow{2}*{\bf LEX} & \multirow{2}*{\bf MV} & \multicolumn{4}{c}{\bf CRF/emb } & \multicolumn{4}{c}{\bf LSTM-CRF/emb } \\ \cmidrule{5-12}
                &     &   &       & {\bf Train} & {\bf +5K} & {\bf +10K} & {\bf +25K} & {\bf Train} & {\bf +5K} & {\bf +10K} & {\bf +25K} \\ \midrule
CDR (Disease)   & 77  & 53.3  & 69.2  & 71.3  & 72.5  &  73.0  & 73.1 &  71.4  & 73.3  & \bf 73.8  &  73.5      \\
CDR (Chemical)  & 28  & 70.3  & 85.7  & 86.2  & 86.9  &  86.9  & 87.1 &  85.2  & 87.2  & 87.3  & \bf 87.4    \\ \midrule
NCBI (Disease)  & 77  & 49.5  & 58.6  & 58.3  & 57.7  &  57.6  & 56.3 &  62.4  & 63.6  & \bf 64.2  &  64.0  \\ 
\bottomrule 
\end{tabular}
\caption{F1 scores for fully automated dictionary supervision, where the only supervision input consists of positive and negative class dictionaries and an optional list of domain-specific stop words. All candidates and labeling functions are generated automatically.}
\label{tbl:fullauto}
\end{table*}

\subsection{Candidate Generation}


\begin{figure}[ht!]
\centering
\includegraphics[width=80mm]{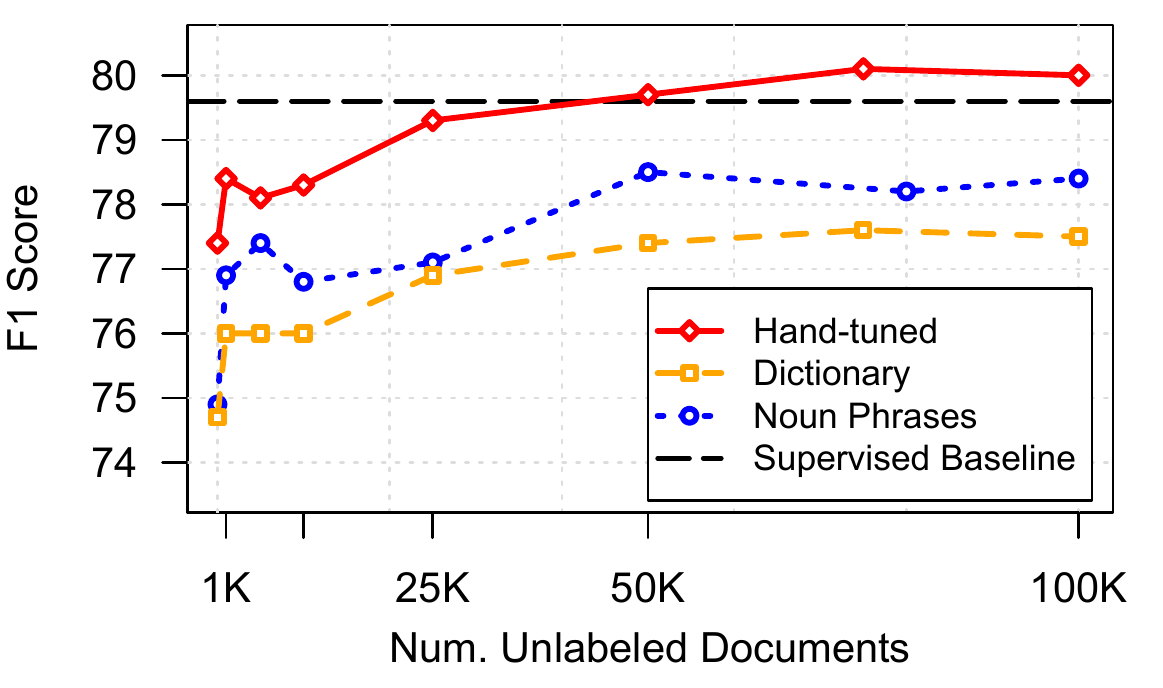}
\caption{Scale-up performance of different candidate generation heuristics on CDR Disease using CRF/emb and extended labeling functions. The hand-tuned heuristic (red) performs better than our supervised baseline (the dashed line).}
\label{fig:cand-gen-tradeoffs}
\end{figure}

Figure \ref{fig:cand-gen-tradeoffs} shows performance trade-offs at scale of our three heuristics in CDR disease with extended labeling functions. Using a hand-tuned candidate generator allows our system to surpass supervised performance, however, completely automated methods also perform very well. A \emph{NounPhrase} generator scores within 2\% of the hand-tuned heuristic with CRF/emb and 1.25\% using  LSTM-CRF/emb (see Table \ref{tbl:dp1}). The bump in performance seen around 1K documents is an artifact of scale-up using a CRF with embeddings, where there is some early overfitting. 

Tables \ref{tbl:dp1} contains scores for PubMed taggers, comparing manual and automated candidate generation heuristics at scale. In both CDR tasks, we can closely match or exceed benchmark scores reported by TaggerOne (from +0.5 to -0.1 F1 points). In chemicals, \emph{NounPhrase} candidate generation demonstrates better recall and improves on hand-tuned matchers by 0.7 points.  In contrast, the NCBI task performs far below baseline using \emph{NounPhrase} generators. This is due to that dataset's more complex definition of mentions, including conjunctions (``breast and ovarian cancer") and prepositional phrases (``deficiency of beta-hexosaminidase A"). These increase partial matches, hurting overall performance. With a hand-tuned candidate generator, we can account for these specific cases and dramatically improve performance and scoring -0.7 F1 points within the benchmark.

\subsection{Autogenerating Supervision}

Table \ref{tbl:fullauto} shows performance measures for our models when using only lexical resources for supervision, without any annotation guideline or dataset-specific labeling functions. In all cases, we find the LSTM-CRF/emb models outperform majority vote by 1.7 to 5.4 F1 points. In chemical tagging, we come within 1 F1 point of published TaggerOne's benchmark score; in NCBI we do much worse due to candidate issues outlined above.

\begin{figure}[ht!]
\includegraphics[width=80mm]{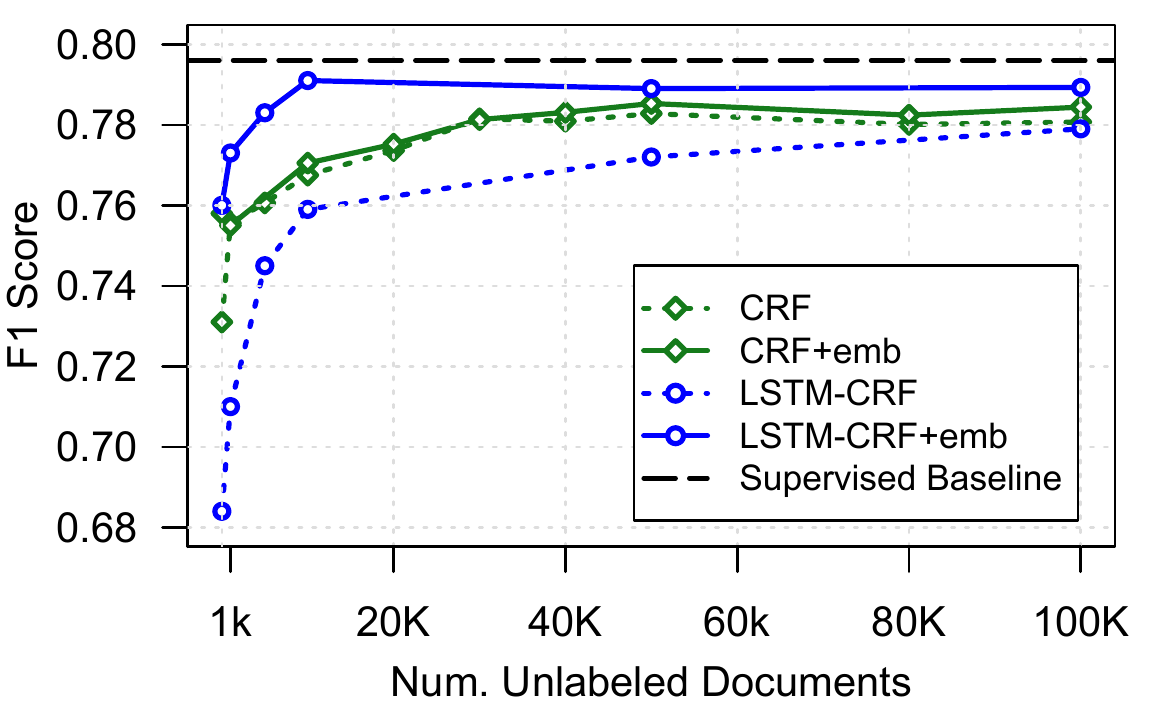}
\caption{Scale-up F1 scores for automatic feature extraction in CDR Disease. The x-axis is the number of unlabeled documents used for training.}
\label{fig:crf-scale}
\end{figure}

\paragraph{Scale-up \& Automatic Feature Extraction:} Figure \ref{fig:crf-scale} gives a broader picture of the scale-up curve and the convergence differences between the human-generated feature library used for our CRF and LSTM-CRF models. Pretrained word embeddings give the LSTM-CRF an advantage in smaller document settings, converging quickly to the best score after 10K additional unlabeled documents. In contrast, without embeddings, the LSTM-CRF is always dominated by CRF models, requiring over 3x more unlabeled data to learn features that approach the same score. Models trained on 100k documents are similar in performance, although the LSTM-CRF is the best overall performer.

\subsection{A Tagger in 24 Hours}

\begin{table}[ht!]
\centering
\setlength\tabcolsep{7pt} 
\begin{tabular}{lcccc}
\toprule
\multirow{2}*{\bf Model} & \multicolumn{4}{c}{\bf i2b2-2009 (Drug)} \\ \cmidrule{2-5}
 &  \bf LFs  & \bf P  & \bf R  &  \bf F1 \\ \midrule
Supervised     &  -  & \bf 90.4 & \bf 88.5 & \bf 89.4 \\ \hline
Auto-LFs MV    & 211 & 84.4 & 62.5 & 71.8 \\
Auto-LFs       & 211 & 82.2 & 74.7 & 78.3 \\ \hline
Custom-LFs MV  & 232 & 90.1 & 69.2 & 78.3 \\
Custom-LFs     & 232 & 83.9 & 82.9 & 83.4 \\ \hline
Lexicon Baseline  &  -  & 31.9 & 67.6 & 43.3 \\
Crowdsourcing  &  79*  & - & - &  $\overline{87.8}$ \\
\bottomrule 
\end{tabular}
\caption{Building a stagger in 24 hours using 1K unlabeled discharge summaries. The crowdsourcing score is the macro-average of 79 annotators.}
\label{tbl:dp3}
\end{table}
Autogenerated labeling functions provide a strong baseline system in this task, scoring 6.5 F1 points over majority vote and boosting recall by 12.2 points. We extending this core system with 21 customized regular expression rules and other guideline specific labeling functions for an overall score within 7\% (6 F1 points) of the same model trained on hand-labeled data. Our approach quickly achieved good performance and, most likely, would improve with more unlabeled training documents which unfortunately are not available for this task.

Comparing our performance to the same task as done with crowdsourcing \cite{uzuner2010community}, we are within 5.1\% (4.4 F1 points) of the crowd macro average achieved by 79 annotators. This required 2 phases of labeling and adjudication over several weeks, although an exact time estimate for drug names alone is difficult as it was one of 1 of 7 annotated subtasks. 

%
%
\section{Conclusion}
\label{sec:conclusion}

In this work, we've demonstrated that programmatic supervision, provided by biomedical lexicons and other heuristics, can achieve competitive performance to state-of-the-art systems trained on hand-labeled data. \systemx accepts much weaker forms of supervision, allowing NER taggers to be built in less time and in a more intuitive fashion for domain experts. Our approach intrinsically scales to automatically construct large training sets, allowing \systemx to train high performance taggers using state of recent deep learning models.

%
%
\section*{Acknowledgments}
This work was supported in part by the Mobilize Center, a National Institutes of Health Big Data to Knowledge (BD2K) Center of Excellence supported through Grant U54EB020405.

\bibliography{emnlp2017.bib}
\bibliographystyle{emnlp_natbib}

%
%
\clearpage
\section{Appendix}
\label{sec:appendix}



\subsection{Materials}

\paragraph{Feature Library:} Our CRF models use feature templates defined over the mention and it's parent dependency parse tree.  Features includes context window features, part of speech tags, word shape, word embeddings, character $n$-grams, morphology, domain dictionary membership, and the lemmatized form of a mention's dependency tree parent word. We expand in-document abbreviations, sharing features across any identical mention linked within a document by a parenthetical mention.

\subsection{Results}

\begin{figure}[ht!]
\centering
\includegraphics[width=80mm]{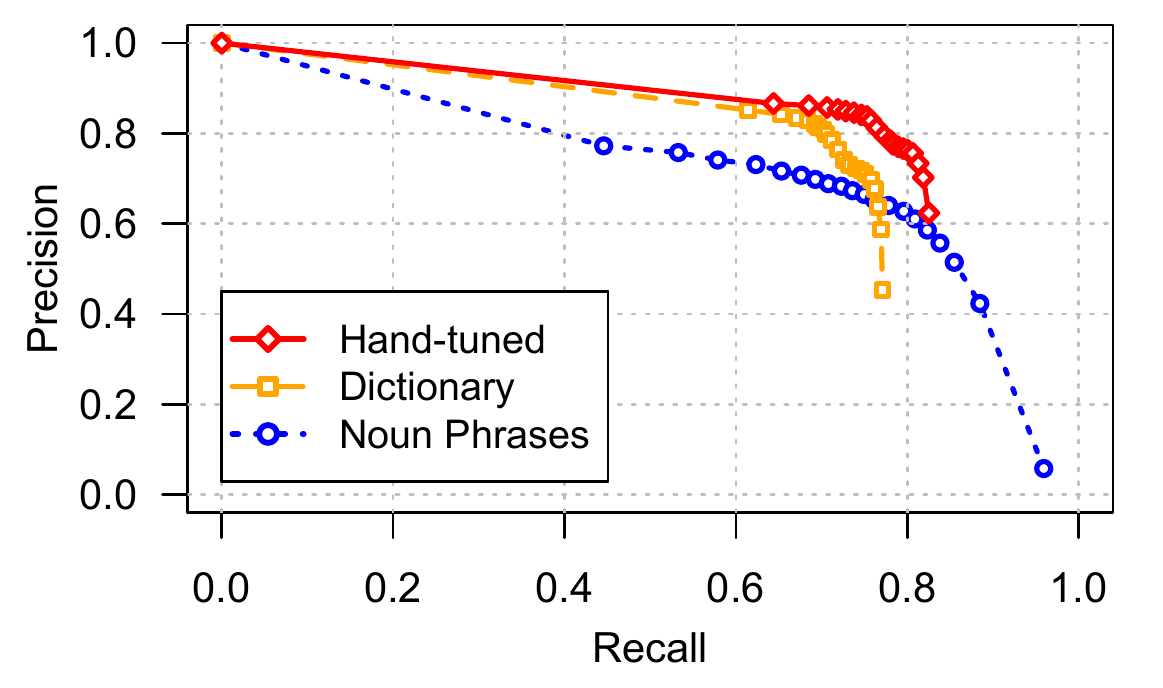}
\caption{Precision-recall curves for candidate generation methods, evaluated on CDR Disease. Phrases of length $<=$4 capture 96\% of all mentions.}
\label{fig:cand-gen-pr}
\end{figure}

\paragraph{Candidate Generation Precision/Recall Curves:} Figure \ref{fig:cand-gen-pr} shows the precision-recall curves of candidate generation methods in detail. Note how domain-engineered matchers suffer recall problems, only generalizing a small amount beyond dictionaries and missing ~20\% of all mentions.


\begin{figure}[ht!]
\centering
\includegraphics[width=80mm]{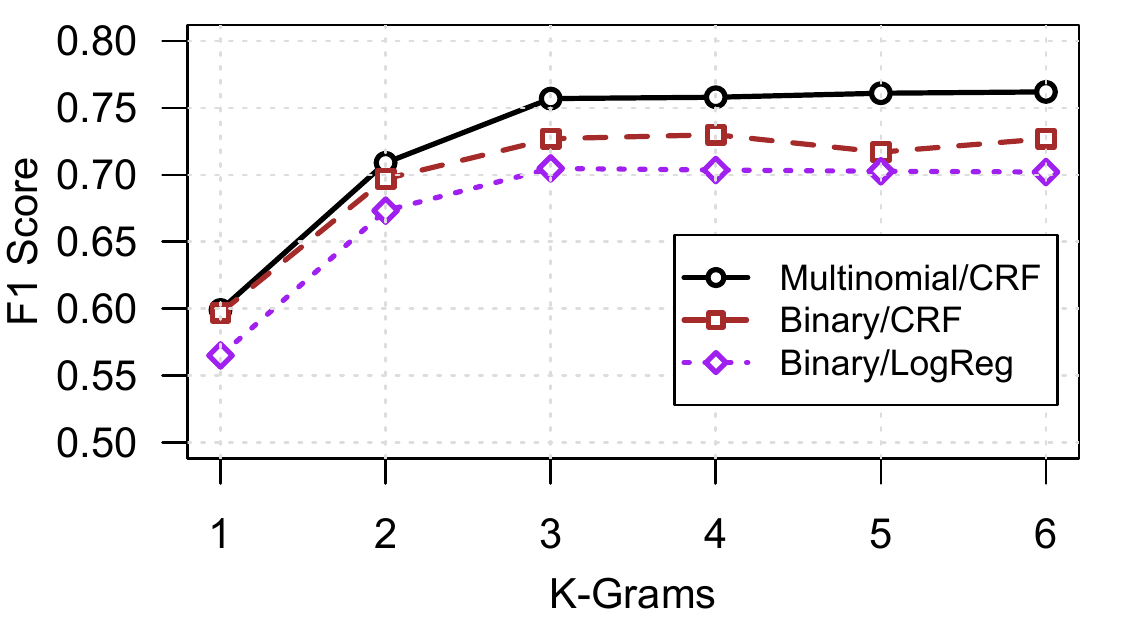}
\caption{F1 scores of the multinomial and binary generative models at different $k$-gram lengths. }
\label{fig:multinomial-tokens}
\end{figure}

\paragraph{Multinomial Generative Model:} Figure \ref{fig:multinomial-tokens} shows the performance difference between the multinomial and binary generative models when sampled data is used to train noise-aware implementations of a CRF and logistic regression. We see the multinomial CRF model performs best overall for all choices of $k$-gram. For sequence models, $k$=6 scored best.

%
\begin{table}[ht!]
\centering
\begin{tabular}{lcccc}
\toprule
\multirow{2}*{\bf Model} & \multicolumn{4}{c}{\bf CDR (Disease)} \\ \cmidrule{2-5}
  &  \bf 500  & \bf +1k  & \bf +10k  &  \bf +100k \\ \midrule
CRF        & 73.1  & 75.6 & 76.8 & 78.0  \\ 
CRF/emb    & 75.8  & 75.5 & 77.1 & 78.4     \\ \midrule
LSTM-CRF       & 68.4 & 71.0 & 75.9 & 77.9 \\ 
LSTM-CRF/emb   & 76.0 & 77.3 & \bf 79.1 & 78.9 \\
\bottomrule 
\end{tabular}
\caption{CDR-Disease scale-up using \emph{NounPhrase} candidates and extended labeling functions.}
\label{tbl:scale-cdr-disease}
\end{table}

%
\begin{table}[ht!]
\centering
\begin{tabular}{lcccc}
\toprule
\multirow{2}*{\bf Model} & \multicolumn{4}{c}{\bf CDR (Chemical)} \\ \cmidrule{2-5}
 &  \bf 500  & \bf +1k  & \bf +10k  &  \bf +25k \\ \midrule
CRF        &  87.5 &  88.0 &  88.1 & 87.2  \\ 
CRF/emb    &  86.7  & 87.7  & 87.9  & 87.2 \\ \midrule
LSTM-CRF       &  78.5  & 79.3  & 82.3  & 85.1 \\ 
LSTM-CRF/emb   &  85.8  & 87.4  & 87.8   & {\bf 88.3} \\
\bottomrule 
\end{tabular}
\caption{CDR-Chemical scale-up using \emph{NounPhrase} candidates.}
\label{tbl:scale-cdr-chemical}
\end{table}

\paragraph{Scaling:} Tables \ref{tbl:scale-cdr-disease} and \ref{tbl:scale-cdr-chemical} show some of the benefits of scale, where we see gains of up to 3.1 F1 points, or 7.6 point improvement over majority vote. Scale-up improvements were smaller in our other tasks, but still beating majority vote in all systems using automatic feature extraction.

\end{document}